\documentclass[journal]{IEEEtran}
\usepackage{amsmath,amsfonts}
\usepackage{algorithmic}
\usepackage{algorithm}
\usepackage{array}
\usepackage[caption=false,font=scriptsize,labelfont=sf,textfont=sf]{subfig}
\usepackage{textcomp}
\usepackage{stfloats}
\usepackage{url}
\usepackage{verbatim}
\usepackage{graphicx}
\usepackage{cite}
\newcommand{\approach}{\textit{TIGR}}

\usepackage{amsmath}

\usepackage{cleveref}

\usepackage{nicefrac}
\usepackage{multirow}

\usepackage{amsthm}
\makeatletter
\def\thm@space@setup{%
  \thm@preskip=0.5em   
  \thm@postskip=0.5em  
}
\makeatother
\newtheorem{definition}{Definition}

\begin{document}
\title{Trajectory Representation Learning on Grids and Road Networks with Spatio-Temporal Dynamics}
\author{Stefan Schestakov,
        Simon Gottschalk \thanks{L3S Research Center, Leibniz University Hannover, 30167 Hannover, Germany, {\tt\small schestakov@L3S.de}, {\tt\small gottschalk@L3S.de}} 
\thanks{}}

\markboth{}%
{}

\maketitle

\begin{abstract}

Trajectory representation learning is a fundamental task for applications in fields including smart city, and urban planning, as it facilitates the utilization of trajectory data 
(e.g., vehicle movements) for various downstream applications, such as trajectory similarity computation or travel time estimation. 
This is achieved by learning low-dimensional representations from high-dimensional and raw trajectory data.
However, existing methods for trajectory representation learning either rely on grid-based or road-based representations, which are inherently different and thus, could lose information contained in the other modality. 
Moreover, these methods overlook the dynamic nature of urban traffic, 
relying on static road network features rather than time-varying traffic patterns. 
In this paper, we propose \approach{}, a novel model designed to integrate grid and road network modalities while incorporating spatio-temporal dynamics to learn rich, general-purpose representations of trajectories. 
We evaluate \approach{} on two real-world datasets and demonstrate the effectiveness of combining both modalities by substantially outperforming state-of-the-art methods, 
i.e., up to 43.22\% for trajectory similarity, up to 16.65\% for travel time estimation, and up to 10.16\% for destination prediction.

\end{abstract}


\section{Introduction}
\label{sec:Intro}

Trajectories are sequences of locations that represent the movements of agents over time. An example of a trajectory is the movement of a vehicle, i.e., a vehicle's position measured at subsequent points in time. 
With the widespread use of location-aware devices, trajectory data is being generated at an unprecedented rate. Due to the importance of mobility as a fundamental aspect of human society and due to abundant trajectory data, there has been a surge in trajectory data analysis, such as travel time estimation \cite{hiereta,auto-stdgcn, rtag}, destination prediction \cite{destination1, destination2, destination3}, and trajectory similarity computation \cite{st2vec, trajgat, kgts}. 

A fundamental task of trajectory analysis is \textit{Trajectory Representation Learning (TRL)}, which aims to learn low-dimensional and meaningful representations of trajectories from high-dimensional and raw trajectory data. 
TRL facilitates the utilization of trajectory data for various downstream applications without the need for manual feature engineering and unique models for specific tasks.

\begin{figure}
\centering
\subfloat[Grid]{%
  \includegraphics[width=0.43\columnwidth]{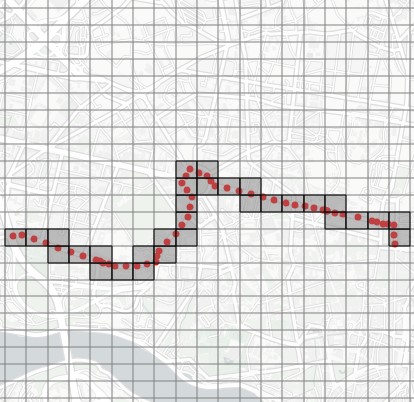}%
  \label{fig:intro:grid}%
}%
\hspace{0.07\columnwidth}%
\subfloat[Road Network]{%
  \includegraphics[width=0.43\columnwidth]{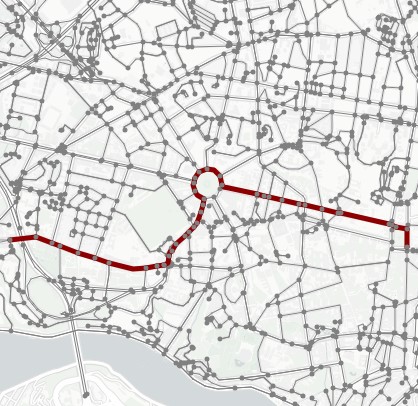}%
  \label{fig:intro:road}%
}%
\caption{Trajectory represented by grid cells (a) and by road segments on a road network (b). © OpenStreetMap, © Carto.}
\label{fig:intro}
\end{figure}

Raw trajectory data often suffers from noise, sparsity, and spatio-temporal irregularities.
TRL methods address these issues using either grid or road network discretization, representing trajectories as sequences of grid cells (Fig. \ref{fig:intro:grid}) or road segments (Fig. \ref{fig:intro:road}). 
Existing TRL methods fall into one of these categories:

\begin{itemize}
    \item \textbf{Grid-based methods} \cite{trajcl,cstte, cltsim, t2vec} divide the spatial domain into a regular grid of equal-sized cells and encode a trajectory as a sequence of these cells. 
    This symmetric, uniform structure is well suited to preserve spatial and structural properties, including shape, distance, and direction.
    \item \textbf{Road-based methods} \cite{start, lightpath, jclm, trembr} represent a trajectory by mapping its locations onto a road network and obtaining a sequence of road segments. 
    Road-based methods inherently model motion restrictions and can exploit knowledge contained in road networks, such as traffic, road features, and topology.  
\end{itemize}

Both modalities are inherently different, thus capturing different trajectory characteristics and possessing unique advantages. 
However, current TRL methods rely solely on one of both modalities and under this restriction could fail to capture vital information on trajectories and lead to inferior representations. 
Moreover, there is a lack of comparative studies between both modalities. 
So far, TRL methods have merely adapted methods of the other modality to their modality by using road-sequences instead of grid-sequences as input and vice versa \cite{start, cstte}. 
Thus, there is an insufficient understanding of each modality's respective advantages.
%

Further, we observe that current TRL methods largely neglect spatio-temporal dynamics. 
Although recent approaches incorporate temporal data, they still treat the road network as static \cite{start, lightpath}. 
This limitation prevents the extraction of crucial spatio-temporal patterns, such as drastically changing traffic conditions during peak hours. 
These dynamics greatly influence downstream applications such as travel time estimation, highlighting the need to incorporate spatio-temporal dynamics in TRL approaches.

In this paper, we propose \approach{} --- a framework for \textbf{T}RL \textbf{I}ntegrating \textbf{G}rid and \textbf{R}oad network modalities with spatio-temporal dynamics. 
%
Based on our previous observations, we identify two main challenges: 
(1) \textit{Heterogeneity between grid and road modality}. Both modalities use fundamentally different spatial discretization, making it challenging to directly combine or fuse both types of information. 
(2) \textit{Spatio-temporal dynamics}. Urban traffic is highly dynamic and varies significantly across time and space. Modeling those dynamics effectively is highly challenging. 

To address these challenges, \approach{} employs a three-branch architecture that processes grid data, road network data, and spatio-temporal dynamics in parallel. 
The grid and road branches represent our two primary modalities, while the spatio-temporal branch extracts dynamic traffic patterns derived from the road modality. 
We embed each branch into latent trajectory representations and align representations within (intra-modal) and across (inter-modal) the distinct branches.  
By combining these three components, \approach{} facilitates a more sophisticated extraction of diverse trajectory characteristics. 
This integrated approach allows our model to leverage the strengths of both grid-based and road-based methods while also accounting for the dynamic nature of urban traffic.

We validate the effectiveness of \approach{} on two real-world datasets and three downstream tasks. 
Our approach outperforms state-of-the-art TRL methods across all settings, demonstrating the effectiveness of combining road and grid modalities for trajectory embeddings.
In summary, the main contributions of this paper are:

\begin{itemize} 
\item We propose \approach{}, a novel TRL model designed to integrate grid and road network modalities, 
outperforming TRL methods across multiple downstream tasks.
\item We develop a novel Spatio-Temporal Extraction method, enabling \approach{} to extract traffic patterns and temporal dynamics.
\item We leverage a three-branch architecture to process grid data, road network data, and spatio-temporal dynamics in parallel and align all three branches through an inter- and intra-modal contrastive loss.
\item To the best of our knowledge, we conduct the first study to compare road-based and grid-based TRL methods across various downstream tasks, revealing that each modality exhibits distinct strengths in different applications.
\item To inspire future research, we release our TRL training and evaluation framework\footnote{The code is available at: \url{https://anonymous.4open.science/r/TRL-9C9D} (will be made public upon acceptance)}, including \approach{} and all evaluated
baselines, data preprocessing, and downstream tasks.
\end{itemize}

\section{Problem Definition}
\label{sec:Problem}

In this section, we first present the notations and then formally define our task.

\begin{definition}
(\textbf{Trajectory}). A trajectory represents the movement of an agent recorded by a GPS-enabled device. 
Formally, a trajectory can be denoted as $\mathcal{T} = [(x_i, y_i, t_i)]_{i=1}^{|\mathcal{T}|}$, where $(x_i, y_i)$ is the location represented as a spatial coordinates pair and $t_i$ is the timestamp of the location at time step $i$. $|\mathcal{T}|$ is the trajectory length and varies for different trajectories.
\end{definition}

Grid-based methods model trajectories by first defining a grid over the spatial domain (e.g., representing a city) and then mapping each point onto its corresponding grid cell.

\begin{definition}
(\textbf{Grid}) A grid is a regular partition of the spatial domain into $M \times N$ cells. Each cell has a unique pair of indices $(m,n)$, where $m \in \{1, \ldots, M\}$ and $n \in \{1, \ldots, N\}$. The set of all grid cells is denoted by $C$, where $C=\{c_1,c_2, \ldots ,c_{|C|}\}$ and each $c_i$ is associated with a unique pair of indices $(m,n)$.
\end{definition}

\begin{definition}
(\textbf{Grid-based Trajectory}). A grid-based trajectory 
$\mathcal{T}^{g} = [(c_i, t_i)]_{i=1}^{|\mathcal{T}^{g}|}$ 
represents a trajectory as a sequence of grid cells, where each $c_i$ denotes a grid cell in the grid $C$.
\end{definition}

Road-based methods model trajectories based on a road network by mapping each point onto the road network using a map matching algorithm \cite{fastmm}.

\begin{definition}
(\textbf{Road Network}) We define a road network as a directed graph $G = (\mathcal{V}, \mathcal{A}, \mathcal{F})$. $\mathcal{V}$ is a set of nodes, where each node $v_i \in \mathcal{V}$ represents a road segment. $\mathcal{A}$ is the adjacency matrix, where $\mathcal{A}_{ij} = 1$ implies that a road segment $v_j$ is directly accessible from road segment $v_i$, and $\mathcal{A}_{ij} = 0$ otherwise. A road network has a feature set $\mathcal{F} \in \mathbb{R}^{|\mathcal{V}| \times f }$ representing road segment features with dimension $f$.
\end{definition}


\begin{definition}
(\textbf{Road-based Trajectory}). A road-based trajectory $\mathcal{T}^{r} = [(v_i, t_i)]_{i=1}^{|\mathcal{T}^{r}|}$
represents a trajectory constrained by a road network $G$, where each $v_i \in \mathcal{V}$ denotes a road segment in the road network $G = (\mathcal{V}, \mathcal{A}, \mathcal{F})$. 
\end{definition}

\textbf{Problem Statement.} Given 
a set of trajectories $\mathcal{D} =\allowbreak \{\mathcal{T}_i\}_{i=1}^{|\mathcal{D}|}$, the task of trajectory representation learning is to learn a trajectory encoder $F: \mathcal{T} \mapsto \mathbf{z}$ that embeds a trajectory $\mathcal{T}$ into a general-purpose representation $\mathbf{z} \in \mathbb{R}^{d} $ with dimension $d$. The representation $\mathbf{z}$ should accurately represent $\mathcal{T}$ suitable for various downstream applications, such as trajectory similarity computation, travel time estimation, and destination prediction. To this end, the trajectory encoder is trained in an unsupervised fashion.


\section{\approach{} Approach}
\label{sec:Approach}

In this section, we present \approach{}'s training procedure, which aims to produce similar representations between two masked views of the same data instance and between the same data instances across different modalities.  

\begin{figure}[t]
\centering
  \includegraphics[width=0.5\textwidth]{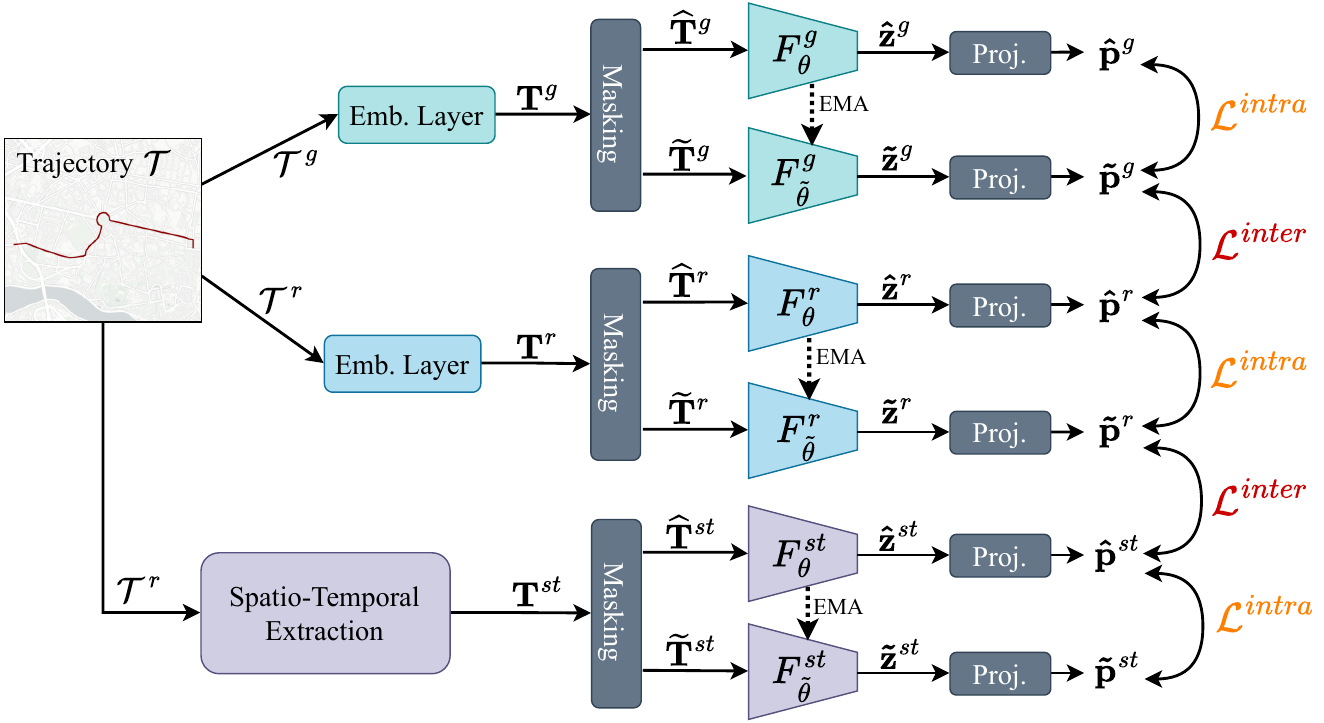}
  \caption{Framework of \approach{}'s training procedure.}
  \label{fig:framework}
\end{figure}

As depicted in Fig. \ref{fig:framework},
\approach{} integrates two primary modalities, grid and road, along with a dedicated spatio-temporal component derived from the road modality. 
Thus, our architecture consists of three parallel branches processing grid, road network, and spatio-temporal dynamics respectively. 
Each branch embeds a trajectory $\mathcal{T}$ into a sequence of token embeddings, thus obtaining $\mathbf{T}^{g}$, 
$\mathbf{T}^{r}$ and $\mathbf{T}^{st}$. 
We leverage distinct embedding layers for $\mathbf{T}^{g}$ and $\mathbf{T}^{r}$ to capture the structural properties of both modalities. 
The spatio-temporal component, on the other hand, aims to extract traffic patterns and temporal dynamics and is detailed in the next section. 
Subsequently, we apply masking to each branch to obtain distinct views and embed each view of each branch using a modality-agnostic encoder. 
The objective is then to align the representations within each branch (intra-modal loss) and across branches (inter-modal loss). 


\subsection{Spatio-Temporal Extraction}
\label{sec:Approach:timeembs}

Modeling spatio-temporal dynamics is crucial for effective TRL. 
However, current methods neglect temporal information completely \cite{lightpath, cltsim}, 
or include temporal information, but solely model static road features \cite{start, mmtec}. 
Thus, these approaches fail to capture the dynamic nature of traffic patterns changing over time. To address this limitation, we propose a novel spatio-temporal extraction method that effectively models both --- dynamic traffic patterns and temporal regularities. 

Our approach consists of three main components: dynamic traffic embedding, temporal embedding, and fusion via local multi-head attention. 
First, we develop a dynamic traffic embedding that captures changing traffic conditions across different times of day, using graph convolutions over transition probabilities to model traffic flow patterns. 
Second, we introduce a temporal embedding that allows our model to capture recurring temporal patterns (e.g., daily rush hours, weekly trends) and temporal dependencies.
Finally, we fuse those embeddings by proposing the local multi-head attention (LMA), based on the observation that traffic within a small spatial and temporal window tends to be very similar, while traffic on distant roads or at different times can differ significantly. 
Our idea is to leverage the multiple heads used in multi-head attention to perform local attention over short sequences. 
This approach of spatio-temporal extraction enables \approach{} to learn rich representations that account for the complex, time-varying nature of urban traffic.

\subsubsection{Dynamic Traffic Embedding} 
To capture dynamic traffic behavior and traffic flow, we leverage graph convolutions over transition probabilities. 
Let $X \in \mathcal{R}^{|t_h|}$ be the aggregated traffic matrix, containing mean traffic speeds for each hour $t_h$.
Further, $P \in \mathcal{R}^{|\mathcal{V}| \times |\mathcal{V}|}$ is the transition probability matrix between connected road segments based on the frequency of trajectories passing through them. 
For two connected road segments $v_i$ and $v_j$, the transition probability is given by $P_{[i,j]}$ and is calculated from historical trajectories, 

\begin{equation}
\label{eq:transition_mx}
P_{[i,j]} =  \frac{ \#transitions (v_i \rightarrow v_j) + 1}{\#total\_visits (v_i) + |\mathcal{N}(v_i)|},
\end{equation}
where $|\mathcal{N}(v_i)|$ is the number of $v_i$'s neighbors.
To leverage $P$ for graph convolutions, we further add self-loops and normalize, i.e., $P'= D^{-1}(P+I)$, where $D$ is the degree matrix and $I$ the identity matrix. 
Formally, the transition probability weighted graph convolution is then defined by:

\begin{equation}
    \mathbf{h}_i = P'_{[i,i]} w_i \mathbf{x}^{(t_d, t_h)}_i  + \sum_{v_j \in \mathcal{N}(v_i)}  P'_{[i,j]} w_j \mathbf{x}^{t_h}_j , 
\end{equation}
where $\mathcal{N}$ is the neighbor function returning all neighbors of a road, 
$w$'s are learnable parameters, $\mathbf{x}^{t_h} \in X$ is the traffic state at hour $t_h$, and  $\mathbf{h}_i$ is the dynamic traffic embedding of road segment $v_i$. 
%
Given the sequence of roads in $\mathcal{T}^{r}$, we obtain the sequence of dynamic traffic embeddings $\mathbf{T}^s = (\mathbf{h}_1, \mathbf{h}_2, \mathbf{h}_3, \dots, \mathbf{h}_{|\mathcal{T}^{r}|})$.

\subsubsection{Temporal Embedding} 
Inspired by the sinusoidal Positional Embeddings \cite{transformer}, we 
learn time embeddings by utilizing the cosine function to capture periodic behavior. Let $t_i$ be the $i$-th time point in a trajectory, then we obtain its time embedding $\mathbf{t}_i$: 
\begin{equation}
    \mathbf{t}_i[k] = \begin{cases} w_k t_i + \phi_k & \text{if } k = 0, \\ \cos(w_k t_i + \phi_k) & \text{else}, \end{cases}
\end{equation}
where $\mathbf{t}_i[k]$ is the $k$-th element of the embedding vector $\mathbf{t}_i \in \mathbb{R}^q$ with dimension $q$, and $w_k$ and $\phi_k$ are learnable parameters.
This representation of time allows periodic behaviors to be captured through the cosine activation function and non-periodic patterns through the linear term. 
Then, we define the sequence of temporal embeddings of a trajectory $\mathcal{T}^{r}$ as $\mathbf{T}^t = (\mathbf{t}_1, \mathbf{t}_2, \mathbf{t}_3, \dots, \mathbf{t}_{|\mathcal{T}^{r}|})$.

\subsubsection{Fusion via Local Multi-Head Attention}
To fuse the dynamic traffic embeddings $\mathbf{T}^s$ and the temporal embeddings $\mathbf{T}^t$, we leverage multi-head attention with a specific adaptation based on the intuition that traffic within a small spatial and temporal window tends to be very similar, while traffic on distant roads or at different times can differ significantly.
Thus, we do not want the attention mechanism to attend to the whole sequence. 
Instead, we leverage each individual head used in multi-head attention to perform local attention over a specific subsequence, proposing the local multi-head attention.

Let $\mathbf{T} \in \mathbb{R}^{|\mathbf{T}| \times d}$ be the input with sequence length $|\mathbf{T}|$ and dimension $d$.
For each head $h \in {1, ..., H}$, we split $\mathbf{T}$ into $H$ subsequences along the sequence dimension, i.e., $\mathbf{T}_h \in \mathbb{R}^{|\mathbf{T}|/H \times d}$, 
and project each $\mathbf{T}_h$ into query, key and value matrices:

\begin{equation} 
Q_h = \mathbf{T}_h W^Q_h, \ 
K_h = \mathbf{T}_h W^K_h, \ 
V_h = \mathbf{T}_h W^V_h,
\end{equation}

where $W^Q_h$, $W^K_h$, $W^V_h \in \mathbb{R}^{d \times d}$ are the weight matrices.
Next, we apply the standard attention mechanism for each head:

\begin{equation}
\label{eq:att}
Att_h(Q_h, K_h, V_h) =  softmax( \frac{Q_h K_h^T}{\sqrt{d}}) V_h .
\end{equation}

Finally, we define the local multi-head attention (LMA) as the concatenation of each attention head along the sequence dimension and a projection with the weight matrix $W^O$:
\begin{equation}
\label{eq:mha}
LMA(Q, K, V) = [ Att_1 || Att_2 || \dots || Att_{|H|} ] W^O.
\end{equation}

To fuse the dynamic traffic and temporal embeddings, we leverage cross-attention utilizing the previously proposed LMA and a concatenation operation:

\begin{equation}
    \mathbf{T}^{st} = LMA( \mathbf{T}^s,\mathbf{T}^t,\mathbf{T}^t )  || LMA( \mathbf{T}^t, \mathbf{T}^s, \mathbf{T}^s ).
\end{equation}

This allows both embeddings to attend to the other and to learn more intricate correlations between dynamic traffic and time periodicity through local attention.


\subsection{Masking}
\label{sec:Approach:augs}

After embedding each branch, we obtain a sequence of token embeddings $\mathbf{T}^{b}$ for each of the branches $b \in \{g, r, st\}$. 
Next, we apply masking by randomly dropping some of the tokens and obtain two distinct views for each branch: $\mathbf{\widehat{T}}^{b}$ (View 1) and  $\mathbf{\widetilde{T}}^{b}$(View 2).
The role of masking is twofold. 
First, we obtain two different views of the same data instance to guide the contrastive learning within each branch. 
Second, similar to how masked autoencoders (MAE) \cite{mae} learn representations, 
the encoder has to develop an understanding of spatial relationships and learn intrinsic information about the trajectory, i.e., the underlying route. 
We investigate three masking strategies:

\begin{itemize}
    \item \textbf{Random masking (RM):} We randomly select a subset of tokens in a trajectory $\mathbf{T^m}$ and mask (i.e., drop) them. We drop each point with a probability of $p_{RM} \in (0,1)$.
    While this is the most commonly used masking strategy across various domains \cite{mae_st, mavil, bert}, inferring a road segment given its adjacent road segments may be trivial.    
    \item  \textbf{Consecutive masking (CM):} A more challenging augmentation is to mask consecutive points, as this forces the model to learn the most probable routes. 
    We select a consecutive amount of points within the trajectory, where the amount of points is defined by the share $p_{CM} \in (0,1)$ of the length of the trajectory $|\mathbf{T^m}|$.
    \item  \textbf{Truncation (TC):} Similarly, we can mask consecutive points from the origin or destination of a trajectory $\mathbf{T^m}$, leading to a truncation of the trajectory.  
    The parameter $p_{TC} \in (0,1)$ defines the share of points to be masked. 
\end{itemize}

In contrast to prior work, we stack multiple masking strategies, yielding a more challenging pretext task and more effective representations, as demonstrated by our experiments (Section \ref{sec:ev:augs}).



\subsection{Encoder}
\label{sec:Approach:enc}

The encoder transforms each masked trajectory into a low-dimensional embedding. 
Our approach employs two types of encoders: 
a target encoder $F^b_{\tilde{\theta}}(\cdot)$ that encodes $\mathbf{\widetilde{T}}^{b}$ into $\mathbf{\tilde{z}}^{b} \in \mathbf{R}^d$, 
and an anchor encoder $F^b_{{\theta}}(\cdot)$ that encodes $\mathbf{\widehat{T}}^{b}$ into $\mathbf{\hat{z}}^{b} \in \mathbf{R}^d$. 
Both encoders have different parameters, where the target encoder's parameters $\tilde{\theta}$ are updated using an exponential moving average (EMA) of the anchor encoder's parameters $\theta$:
$\tilde{\theta} \leftarrow \mu \tilde{\theta} + (1-\mu) \theta$, 
where $\mu \in [0,1]$ is a target decay rate. This EMA update ensures a more stable learning target and prevents representation collapse \cite{msn}.

For the encoder architecture, we base our design on the Transformer \cite{transformer} with specific enhancements. 
We incorporate Rotary Positional Embeddings (RoPE) \cite{roformer}, which capture relative positional information during the attention process, improving upon standard positional embeddings that only represent absolute positions. 
Additionally, we utilize Root Mean Square Layer Normalization (RMSNorm) \cite{rmsnorm} and apply it before the multi-head attention and feed-forward layer rather than after. 
This modification further stabilizes the training process. 
To obtain the final trajectory representation from each branch, we apply mean pooling over the output sequence of the transformer's final layer.
These architectural choices enable our encoders to effectively capture the complex spatial and temporal relationships within trajectory data.


\subsection{Inter- and Intra-Modal Losses}
\label{sec:Approach:train}

Next, we project each representation through a \textit{projection head}, modeled as an MLP. 
The use of projection heads has been reported to increase performance in contrastive learning \cite{nnclr, cltsim}. 
After projection of $\mathbf{\tilde{z}}^{b}$ and $\mathbf{\hat{z}}^{b}$, we obtain  $\mathbf{\tilde{p}}^{b}$ and $\mathbf{\hat{p}}^{b}$.
We propose to employ the contrastive objective within each branch (intra-modal) and across branches (inter-modal) to learn rich and diverse representations that capture complementary information from both modalities and the spatio-temporal component. 
For the contrastive objective, we leverage the InfoNCE loss \cite{InfoNCE}:

\begin{equation}
    \label{eq:nceloss}
    \mathcal{L}(x,y) = -\log \frac{\exp(x \cdot y / \tau)}{\sum_{j=0}^{|\mathcal{Q}_{neg}|} \exp(x \cdot y_j / \tau) +  \exp(x \cdot y / \tau)},
\end{equation}
where $\tau$ is the temperature parameter, and $\mathcal{Q}_{neg}$ is the negative sample queue, updated at each iteration to maintain a large set of negative samples.
At each iteration, we enqueue the current batch of samples and dequeue the oldest batch of samples from the queue.
Then we align views within each branch (intra-modal) and across distinct branches (inter-modal):


\textbf{Intra-modal} contrastive learning 
enables alignment \textit{within} each branch. 
We contrast the projected embeddings within each branch leveraging the contrastive loss from Eq.~\ref{eq:nceloss}:
\begin{equation}
    \mathcal{L}^{intra} = \nicefrac{1}{3} \sum_{\substack{m \in \\ \{g, r, st\}}} \mathcal{L}(\mathbf{\tilde{p}}^{b}, \mathbf{\hat{p}}^{b}).
\end{equation}


\textbf{Inter-modal} contrastive learning 
facilitates alignment \textit{across} each branch.
We contrast the projections of both structural branches, i.e., $\mathbf{\tilde{p}}^{g}$ and $ \mathbf{\hat{p}}^{r}$, as well as both road network-modality based branches\footnote{The spatio-temporal branch ($st$) is based on road network traffic information, and thus, is very distinct from the grid modality ($g$). Aligning those two branches empirically degraded the quality of learned trajectory representations.}, i.e., $\mathbf{\tilde{p}}^{r}$ and $ \mathbf{\hat{p}}^{st}$.
We define the inter-modal loss as:
\begin{equation}
    \mathcal{L}^{inter} = \nicefrac{1}{2} [\mathcal{L}(\mathbf{\tilde{p}}^{g},\mathbf{\hat{p}}^{r}) +  \mathcal{L}(\mathbf{\tilde{p}}^{r}, \mathbf{\hat{p}}^{st})].
\end{equation}

These contrastive losses promote the alignment 
within and across branches. 
The overall loss function of \approach{} is the weighted sum of the inter and intra loss, with $\lambda$ being a weighting coefficient:
\begin{equation}
\label{eq:losses}
    \mathcal{L}_{\approach{}} =  \lambda \mathcal{L}^{intra} + (1-\lambda) \mathcal{L}^{inter}.
\end{equation}

\subsection{Downstream Application}

%
%

After training, to obtain the final representation $\mathbf{z}$ of a trajectory, we encode the embedding sequence of each branch 
using the trained trajectory encoder $F^b_{\theta}(\cdot)$ and discard the masking and projection heads: 

\begin{equation} 
\mathbf{z}^{g} = \mathrm{F^g_{\theta}}(\mathbf{T}^{g}), \ 
\mathbf{z}^{r} = \mathrm{F^r_{\theta}}(\mathbf{T}^{r}), \ 
\mathbf{z}^{st} = \mathrm{F^{st}_{\theta}}(\mathbf{T}^{st}).
\end{equation}

While we aligned branches through the inter-contrastive loss, the embeddings contain their own specific features. 
Thus, we concatenate all representations to obtain the final trajectory representation
$\mathbf{z} \in \mathbb{R}^{d}$:
\begin{equation} 
\mathbf{z} = \mathbf{z}^{g} || \mathbf{z}^{r}|| \mathbf{z}^{st}.
\end{equation}

\section{Evaluation Setup}
\label{sec:eval-setup}

In this section, we describe the real-world trajectory datasets, baseline methods, and downstream tasks used to evaluate \approach{}'s performance and demonstrate its effectiveness in trajectory representation learning.

\subsection{Datasets}

We conduct our experiments on openly available trajectory and road network datasets of two cities, namely Porto\footnote{\url{https://www.kaggle.com/competitions/pkdd-15-taxi-trip-time-prediction-ii}}, containing 1.6 million taxi trajectories from July 2013 to June 2014,
and San Francisco\footnote{\url{https://ieee-dataport.org/open-access/crawdad-epflmobility}}, containing 0.6 million trajectories from May 2008 to June 2008.
The corresponding road networks are extracted from OpenStreetMap\footnote{\url{https://www.openstreetmap.org/}}. 
We preprocess the trajectory data by pruning trajectories outside the respective city's bounding box, removing trajectories containing less than 20 points or more than 200 points, and matching the trajectories to the road network via Fast Map Matching \cite{fastmm}.

\subsection{Baselines}

As baselines, we select several state-of-the-art trajectory representation learning methods that are trained in a self-supervised manner and can be used for various downstream tasks. 
Thus, we also include those methods proposed for trajectory similarity learning that adopt unsupervised-training, while not considering those methods that are trained supervised, e.g., Traj2SimVec \cite{traj2simvec}, T3S \cite{t3s} and TrajGAT  \cite{trajgat}, or need additional labels, e.g., WSCCL \cite{wsccl}. 
Further, we divide the baselines into two categories, grid-based and road network-based:

\textbf{Grid-based methods}
\begin{itemize}
    \item t2vec \cite{t2vec} uses skip-gram to encode grid cells based on neighbor contexts and trains an LSTM encoder via seq2seq reconstruction of downsampled trajectories.
    \item CLT-Sim \cite{cltsim} pre-trains cell embeddings using skip-gram on cell sequences. CLT-Sim encodes trajectories with LSTM and trains via contrastive learning using NT-Xent loss.
    \item TrajCL \cite{trajcl} pre-trains cell embeddings using node2vec by treating the grid as a graph. Further, TrajCL employs a transformer encoder, which is trained contrastively.
    \item CSTTE \cite{cstte} learns cell embeddings during training without pre-training and adds temporal regularities. The model uses a transformer architecture with contrastive learning and NT-Xent loss.
\end{itemize}

\textbf{Road network-based methods}

\begin{itemize}
    \item Trembr \cite{trembr} pre-trains road embeddings by predicting co-occurrence and road type. The model uses a seq2seq architecture to learn representations through trajectory reconstruction and travel time prediction.
    \item Toast \cite{toast} pre-trains road embeddings using modified deepwalk \cite{deepwalk} with road type prediction. The model employs a transformer architecture trained with a masking objective.
    \item JCLRNT \cite{jclm} learns road representations and trajectory representations in parallel during training. The model applies contrastive learning between roads, trajectories, and road-trajectory pairs.
    \item LightPath \cite{lightpath} aims to reduce model complexity, by utilizing sparse paths and a distillation approach. 
    \item START \cite{start} encodes roads using GAT with transfer probabilities and models temporal regularities. The model uses a transformer encoder trained via contrastive learning and sequence masking.
\end{itemize}

Additionally, to compare road and grid modality, independently of architecture choices, we add a Transformer baseline that uses contrastive learning with a vanilla Transformer for both modalities (TF$^r$ and TF$^g$) and feed ether grid-embeddings or road-embeddings.

\subsection{Downstream Tasks}

We evaluate the performance of TRL methods on three downstream tasks:

\begin{itemize}
    \item \textbf{Trajectory similarity (TS)} aims to find the most similar trajectory in a database $D$ given a query trajectory $T_q$.
    To create the \textit{query set} $Q$ and \textit{database} $D$, we follow Trembr \cite{trembr} and TrajCL \cite{trajcl}. 
    However, in contrast to these, we increase task difficulty, by extending the sample size to $10,000$ trajectories. 
    Further, we split them into two sub-trajectories by sampling the odd points of a trajectory $T^q_{odd}$ and its even points $T^q_{even}$.
    We insert $T^q_{odd}$ into $Q$ and $T^q_{even}$ into $D$.
    To increase the database size, we randomly sample another $k_{neg} = 10,000$ trajectories and add their even samples to $D$.
    For evaluation, we take the dot product of a query trajectory $T_q \in Q$ and each trajectory $T \in D$ and rank them by their similarity. Ideally, the even trajectory corresponding to $T_q$ ranks first. 
    We report mean rank ($MR$) and hit ratio ($HR@1$ and $HR@5$).
    \item \textbf{Travel time estimation (TTE)} predicts the travel time of a trajectory based on its coordinate pairs. It can be used for traffic management, congestion avoidance, or ride-hailing services. We embed each trajectory and train an MLP to predict the travel time in seconds.
    To avoid information leakage, we only feed the start time and mask all other timestamps for this task. 
    We report the mean absolute error ($MAE$), mean absolute percentage error ($MAPE$), and root mean squared error ($RMSE$).
    \item \textbf{Destination prediction (DP)} predicts the final destination of a partially observed trajectory.
    It facilitates route planning, navigation, and recommendation. We embed $90\%$ of each trajectory and train an MLP to predict the final location on the road network. 
    We report the F$_1$-score ($F_1$) and accuracy ($Acc@1$ and $Acc@5$).
\end{itemize}

While some works fine-tune the complete TRL models when training for a downstream task, we follow those that freeze the models and only train a prediction MLP \cite{byol, simclr, trembr, jclm}.

\subsection{Implementation Details}
We train \approach{} using the Adam optimizer with an initial learning rate of $0.001$, a batch size of $512$, and $10$ training epochs.
Based on our masking analysis in Section \ref{sec:ev:augs}, we select the following
masking variants: \textit{truncation} and \textit{consecutive masking} for the first set of masking strategies (View 1) and all three for the other set of masking strategies (View 2). 
We set their parameters to $p_{RM} = p_{TC} = p_{CM} = 0.3$.
Dropout is set to $0.1$ and the temperature parameter $\tau$ to $0.05$.
Based on our parameter analysis in Section \ref{sec:ev:param} we further
use two encoder layers, a negative sample queue of $2,048$, and an embedding dimension of $512$.
We embed grid cells with a dimension of $256$, and road and spatio-temporal branches with a dimension of $128$. 
Similar to previous works \cite{lightpath, trajcl}, we employ node2vec \cite{node2vec} to initialize the grid and road embedding layers. 
Finally, to balance the loss in Eq.~\ref{eq:losses}, we set $\lambda = 0.5$.

\begin{table*}[ht]
    \caption{Overall performance of \approach{} and baselines. Arrows indicate whether higher values ($\uparrow$) or lower values ($\downarrow$) are better. Bold results are the best, and underlined results are the second best.}
     \centering
     \renewcommand*{\arraystretch}{1.1}
      \resizebox{1\textwidth}{!}{
     \begin{tabular}{ | l | l | l | c  c  c | c  c  c | c  c  c | }
     \hline
     \parbox[t]{2.5mm}{} & \parbox[t]{2.5mm}{} & \multicolumn{1}{l|}{Task}  &          \multicolumn{3}{c|}{\textbf{Trajectory Similarity (TS)}} & 
                            \multicolumn{3}{c|}{\textbf{Travel Time Estimation (TTE)}} &
                            \multicolumn{3}{c|}{\textbf{Destination Prediction (DP)}}  \\
    \cline{3-12}
    & & \multicolumn{1}{l|}{Metric} &
                MR $\downarrow$ & HR@1 $\uparrow$ & HR@5 $\uparrow$ &
                MAE $\downarrow$ & MAPE $\downarrow$ & RMSE $\downarrow$ &
                F$_1$ $\uparrow$ & Acc@1 $\uparrow$ & Acc@5 $\uparrow$  \\

    \hline
    \hline
    \parbox[t]{2.5mm}{\multirow{12}{*}{\rotatebox[origin=c]{90}{Porto}}} 
    & \parbox[t]{2.5mm}{\multirow{6}{*}{\rotatebox[origin=c]{90}{Road}}} 
    & TF$^r$ & 6.308 & 0.267 & 0.696 & 149.55 & 23.07 & 219.57 & 0.084 & 0.173  & 0.387   \\
    & & Trembr  & 32.46 & 0.109 & 0.280 & 164.50 & 25.69 & 238.26 & 0.088 & 0.180  & 0.416  \\
    & & Toast & 12.50 &  0.308 & 0.642 & 159.35 & 24.97 &  235.47 & \underline{0.115} & \underline{0.223}  & \underline{0.486}  \\
    & & JCLRNT & 1.691 & 0.775 & 0.958 & 145.05 & 22.01 & 215.21 & 0.085 & 0.176  & 0.410 \\
    & & LightPath & 1.958 & 0.734 & 0.943 & 157.72 & 23.94 & 231.13 & 0.053 & 0.123  & 0.282 \\
    & & START  & 2.581 & \underline{0.816} & 0.948 & 131.40 &  19.94 & 196.71 & 0.074 & 0.156  & 0.371 \\
    \cline{2-12}
    & \parbox[t]{2.5mm}{\multirow{5}{*}{\rotatebox[origin=c]{90}{Grid}}} 
    & TF$^g$ & 9.019 & 0.166 & 0.514 & 110.71 & 17.03 & 171.34 & 0.082 & 0.170  & 0.408  \\
    & & t2vec & 20.83 & 0.169 & 0.413 & 115.27 & 17.68 & 172.83 & 0.093 & 0.189  & 0.458 \\
    & & CLT-Sim & 15.87 & 0.211 & 0.503 & 162.62 & 25.30 & 241.67 & 0.098 & 0.196  & 0.465  \\ 
    & & CSTTE & 3.838 & 0.751 & 0.902 & 113.29 & 17.04 & 175.52 & 0.082 & 0.170  & 0.418  \\
    & & TrajCL & \underline{1.518} & 0.804 & \underline{0.976} & \underline{97.42} & \underline{14.38} & \underline{154.32} & 0.077 & 0.163  & 0.393  \\
    \cline{2-12}
    & & \approach{} & \textbf{1.034} & \textbf{0.976} & \textbf{1.000} & \textbf{86.86} & \textbf{12.76} & \textbf{139.80} & \textbf{0.128} &  \textbf{0.241}  & \textbf{0.527}  \\
    & & \textit{Improvement} & 31.88\% & 19.61\% & 4.06\% & 10.84\% & 11.27\% & 9.41\% & 10.16\% & 8.07\% & 8.44\%  \\
    \hline
    \hline
    \parbox[t]{2.5mm}{\multirow{12}{*}{\rotatebox[origin=c]{90}{San Francisco}}} 
    & \parbox[t]{2.5mm}{\multirow{6}{*}{\rotatebox[origin=c]{90}{Road}}} 
    & TF$^r$ & 5.654 & 0.326 & 0.765 & 121.48 & 22.75 & 188.62 & 0.040 & 0.081  & 0.221   \\
    & & Trembr  & 29.62 & 0.118 & 0.259 & 123.26 & 23.38 & 191.10 & 0.041 & 0.084  & 0.221  \\
    & & Toast & 16.95 & 0.206 & 0.481 & 120.58 & 22.17 & 187.09 & 0.048 & 0.098  & 0.272  \\
    & & JCLRNT & \underline{1.874} & 0.732 & \underline{0.949} & 107.95 & 19.76 & 170.88 & 0.036 & 0.075  & 0.175  \\
    & & LightPath & 2.045 & 0.726 & 0.938 & 118.58 & 22.76 & 181.21 & 0.019 & 0.042  & 0.112 \\
    & & START  &  4.153 & \underline{0.747} & 0.867 & 111.83 & 21.04 & 172.00 & 0.030 & 0.061  & 0.175  \\
    \cline{2-12}
    & \parbox[t]{2.5mm}{\multirow{5}{*}{\rotatebox[origin=c]{90}{Grid}}} 
    & TF$^g$ & 12.00 & 0.162 & 0.447 & 81.35 & 14.71 & 134.67 & 0.039 & 0.080  & 0.224  \\
    & & t2vec & 27.62 & 0.126 & 0.329 & 109.56 & 20.63 & 163.19 & 0.050 & 0.101  & \underline{0.279}  \\
    & & CLT-Sim & 57.10 & 0.033 & 0.100 & 112.98 & 22.64 & 173.16 &  \underline{0.051} &  \underline{0.103}  & 0.265 \\
    & & CSTTE & 14.27 & 0.321 & 0.598 & 80.06 & 14.56 & 139.39 & 0.026 & 0.056  & 0.176  \\
    & & TrajCL & 4.447 & 0.673 & 0.866 & \underline{68.41} & \underline{13.03} & \underline{113.02} & 0.032 & 0.067  & 0.199  \\
    \cline{2-12}
    & & \approach{} & \textbf{1.064} &  \textbf{0.952} & \textbf{0.996} & \textbf{60.38} & \textbf{10.86} & \textbf{106.43} & \textbf{0.055} &  \textbf{0.107}  & \textbf{0.286}  \\
    & & \textit{Improvement} & 43.22\% & 27.44\% & 4.95\% & 11.73\% & 16.65\% & 5.83\% & 7.84\% & 3.88\% & 2.51\%  \\
    \hline
    \end{tabular}
    }
     \label{tab:results}
\end{table*}

\section{Evaluation}
\label{sec:Evaluation}

In this section, we aim to validate \approach{}'s effectiveness by comparing its performance against baseline methods and analyzing the distinct contributions of each branch (grid, road, and spatio-temporal), masking strategies, and parameters through comprehensive experiments. 
Moreover, we conduct a systematic comparison between grid-based and road-based methods to understand their distinct strengths in different downstream tasks.

\subsection{Overall Performance}

Table \ref{tab:results} reports the overall performance based on the mean of five different runs with different random seeds.
Our proposed method \approach{} substantially outperforms all baselines for all metrics on the three downstream tasks and both datasets, 
demonstrating the benefits of integrating road-based and grid-based modalities, coupled with the extraction of dynamic spatio-temporal patterns. 
While \approach{} shows substantial improvements in the \textbf{TTE} and \textbf{DP} tasks (up to 16.65\% and 10.16\% respectively), it demonstrates even greater advancements in the \textbf{TS} task, achieving near-perfect results on both datasets with improvements of up to 43.22\%.

\begin{figure}[bt]
    \centering
  \includegraphics[width=0.5\textwidth]{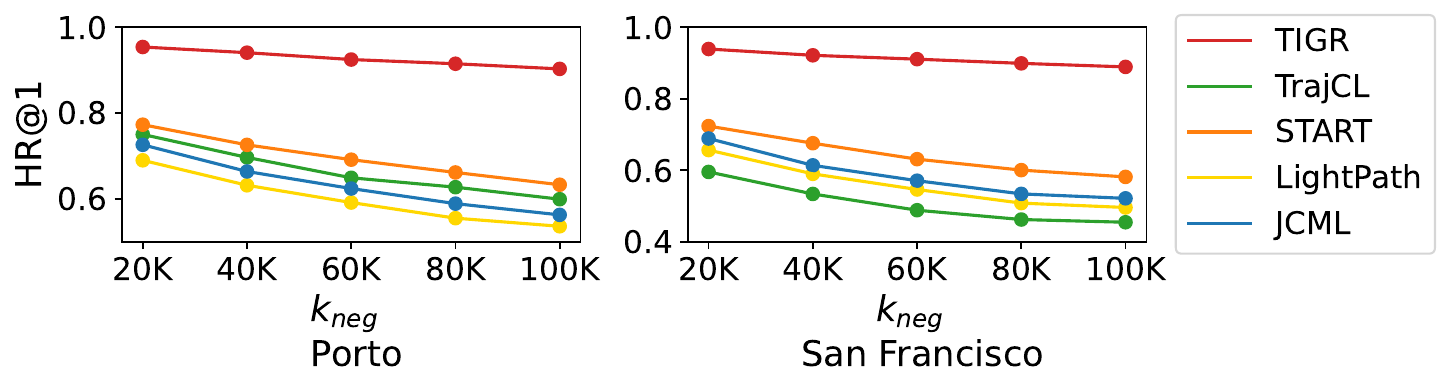}
  \caption{Performance on \textbf{TS} (HR@1 $\uparrow$) with increasing amount of negative samples $k_{neg}$.}
  \label{fig:exp_db}
\end{figure}

To assess robustness in \textbf{TS}, we further increase the task's difficulty by adding more negative samples $k_{neg}$ (as illustrated in Fig. \ref{fig:exp_db}). 
As $k_{neg}$ increases, the performance of all methods declines. 
Nevertheless, \approach{} exhibits a significantly slower rate of decrease, maintaining scores close to 1.0, while the baseline models suffer substantial declines in performance.

These results demonstrate that combining both modalities and including spatio-temporal dynamics substantially benefits all tasks, 
with the most notable improvements observed in \textbf{TS}. 
This suggests that grid- and road-modality capture unique trajectory characteristics, with the integration of both enabling more effective and comprehensive trajectory representations than single-modality approaches.

\subsection{Comparative Analysis of Modalities}

From Table \ref{tab:results}, we first compare modalities based on the Transformer baseline, which uses the same architecture for both modalities. 
On both datasets, road-based TF$^r$ demonstrates superior performance over TF$^g$ in \textbf{TS}, for instance, 30\% improvement in terms of MR on Porto. 
In contrast, grid-based TF$^g$ excels in \textbf{TTE}, substantially improving performance compared to TF$^r$, e.g., 26.2\% improvement in terms of MAPE on Porto. 
This demonstrates that given the same architecture, both modalities encode unique features and excel at different tasks.

Further, we compare baselines of both modalities\footnote{Note that previous works which compared to methods of the other modality, adapted them to their modality \cite{start, jclm}, e.g., using $\mathbf{T}^r$ as input instead of $\mathbf{T}^g$ and vice versa.} and observe similar trends.
For \textbf{TTE}, grid-based methods are clearly superior to road-based methods, with TrajCL beating the best road-based method START by 27.5\% in terms of MAPE. 
Even less sophisticated models, such as t2vec, outperform all road-based methods. 
While JCLRNT and START show improvements by incorporating travel semantics, they only consider static road features and thus cannot capture dynamic spatio-temporal patterns. 
For \textbf{TS}, road-based methods outperform grid-based methods on San Francisco, while on Porto both modalities' methods are competitive.
%
For \textbf{DP}, the performance disparity between modalities is less pronounced, indicating that both modalities capture relevant features for this task.
Our comparative analysis reveals that road-based and grid-based modalities exhibit complementary strengths across different trajectory tasks, emphasizing their respective strengths in capturing distinct aspects of trajectory data.

\subsection{Ablation Study}

In our ablation study, we evaluate the performance of individual branches, their combinations, and the impact of removing various components one at a time. Table \ref{tab:abl} reports the results, demonstrating the unique contributions of each component and branch in \approach{}.

When comparing individual branch performance of grid and road, the road branch outperforms the grid branch in both \textbf{TS} and \textbf{DP}, achieving 4.6\% higher HR@1 and 8.8\% higher Acc@1, respectively.
However, the grid branch excels at \textbf{TTE}, improving MAPE by 41.5\%, again suggesting that the grid modality is crucial for accurate travel time predictions. 
The spatio-temporal branch models dynamic traffic and temporal patterns of roads and thus is slightly ahead of the road branch in terms of \textbf{TTE}, which only models structural properties. 
However, due to the lack of structural information, the spatio-temporal branch fails at \textbf{DP} and achieves lower performance on \textbf{TS} of 0.596 in terms of HR@1.

When combining pairs of branches, we observe complementary effects. 
The combination of both structural branches (grid and road) shows strong performance across all tasks, particularly excelling in \textbf{DP} (Acc@1 of 0.241). 
The addition of the spatio-temporal branch to either structural branch significantly improves performance in \textbf{TS} and \textbf{TTE} (e.g., HR@1 by 23.4\% and MAPE by 18.8\% for the combination with the road branch), demonstrating the impact of effectively modeling spatio-temporal dynamics.
Overall, the combination of branches increases downstream performance, with leveraging all three branches achieving the highest performance overall. 

The removal of architectural components like inter-modal loss, Local Multi-head Attention (LMA), and Rotary Position Embedding (RoPE) each cause slight decreases across tasks, though their impact is less dramatic than removing entire branches, suggesting that these components provide essential but incremental improvements.

The complete \approach{} model, incorporating all branches and components, achieves the best overall performance, demonstrating that each element contributes to the model's effectiveness. 
These results highlight the importance of \approach{}'s multi-branch architecture and validate our approach of combining grid, road, and spatio-temporal information for comprehensive trajectory representation learning.

\begin{figure*}[t]
    \centering
  \includegraphics[width=1\textwidth]{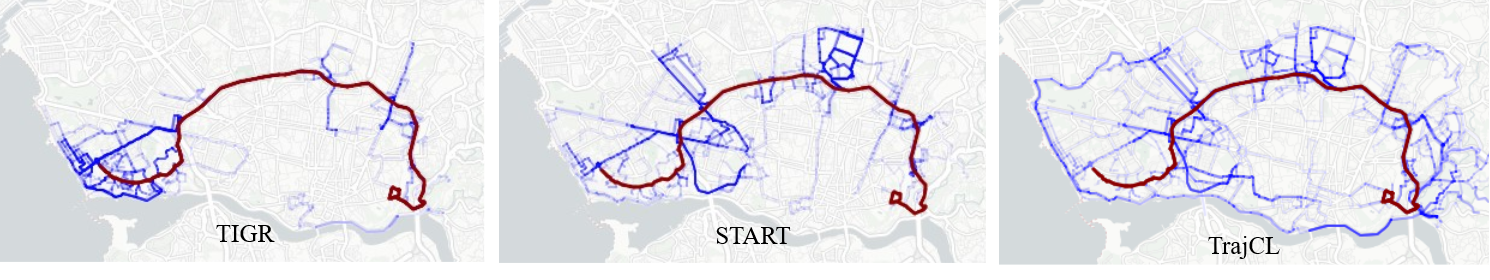}
  \caption{Given the query trajectory (red), we visualize the 500 most similar trajectories retrieved by \approach{} (left), START (middle), and TrajCL (right).}
  \label{fig:exp_500sim}
\end{figure*}

\begin{table}[t]
\centering
\caption{Ablation study on the Porto dataset.}
\begin{tabular}{l|c|c|c}
\hline
\textbf{ } & \textbf{TS} & \textbf{TTE} & \textbf{DP} \\
& HR@1 $\uparrow$ & MAPE $\downarrow$ & Acc@1 $\uparrow$ \\
\hline
w/o Road \& ST (= only Grid) & 0.702 & 13.49 & 0.193 \\
w/o Grid \& ST (= only Road) & 0.734 & 23.05 & 0.210 \\
w/o Road \& Grid (= only ST) & 0.596 & 20.09 & 0.031 \\
w/o ST branch (= Grid \& Road) & 0.849 & 13.33 & 0.241 \\
w/o Road branch (= Grid \& ST) & 0.931 & 13.06 & 0.189 \\
w/o Grid branch (= Road \& ST) & 0.958 & 18.71 & 0.213 \\
w/o inter-loss & 0.973 & 13.22 & 0.214 \\
w/o LMA & 0.963 & 13.30 & 0.236 \\
w/o RoPE & 0.973 & 12.94 & 0.236 \\
\hline
\approach{} & \textbf{0.976} & \textbf{12.76} & \textbf{0.241} \\
\hline
\end{tabular}
\label{tab:abl}
\end{table}

\subsection{Impact of Masking Strategies}
\label{sec:ev:augs}
\begin{figure}[tb]
    \centering
  \includegraphics[width=0.45\textwidth]{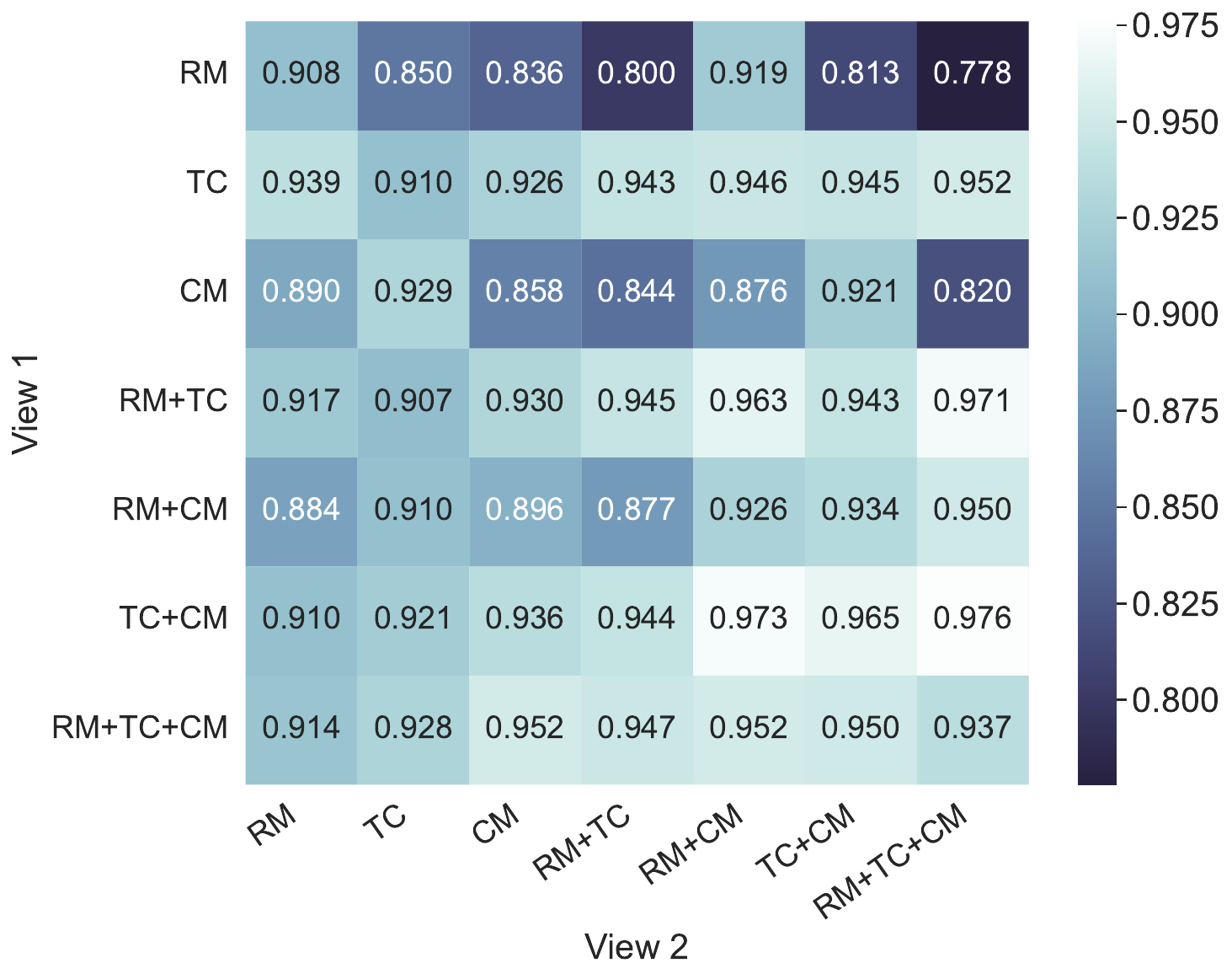}
  \caption{Impact of the masking strategies on the Porto dataset. We report HR@1 ($\uparrow$) of the \textbf{TS} task.}
  \label{fig:exp_augs}
\end{figure}

We study the impact of masking methods by varying masking strategies applied to each view on the \textbf{TS} task on Porto. 
We utilize random masking (RM), truncation (TC), and consecutive masking (CM) and their combinations and report the results in Fig. \ref{fig:exp_augs}.
Masking significantly impacts the downstream performance, with an improvement of 25.4\% between the least and best-performing combinations.
Only the \textit{anchor encoder} processing the trajectories modified by View 1 masking (presented on the y-axis) is trained through gradient descent. 
To this extent, we observe that masking applied to View 1 has a greater impact on performance. 
More specifically, TC is the best individual masking strategy, and combinations containing TC are superior to those without it. 
Further, using only one masking strategy per view, as done by previous contrastive TRL methods \cite{trajcl, start}, is inferior to using a combination of masking strategies for each view. 
Finally, based on the best-performing combination, we select TC and CM for View 1 and all three for View 2.

\subsection{Comparison of Top-500 Similar Trajectories}
\label{sec:ev:top500}

To provide a visual comparison of similar trajectory search results, we present the 500 most similar trajectories retrieved by \approach{}, START, and TrajCL for a given query trajectory in Fig. \ref{fig:exp_500sim}. 
The query trajectory (depicted in red) originates from Porto's main station, traverses the highway, and ends in the coastal area.
\approach{} demonstrates a significantly higher overlap between retrieved trajectories and the query trajectory. 
Minor deviations are primarily observed in the coastal region, suggesting that \approach{} predominantly identifies trajectories that originate from the main station and terminate at various specific locations within the coastal area. 
In contrast, similar trajectories retrieved by START and TrajCL exhibit substantially less overlap with the query trajectory. 
While these trajectories generally maintain connectivity with the query path, they cover different areas along the route. 
These findings further illustrate that the integration of grid and road modalities enhances the modeling of intricate and diverse trajectory characteristics, thereby facilitating the retrieval of highly similar trajectories.

\subsection{Parameter Study}
\label{sec:ev:param}
\begin{figure}[t]
    \centering
    
    \subfloat[Amount of encoder layers.]
    {\includegraphics[width=0.5\textwidth]{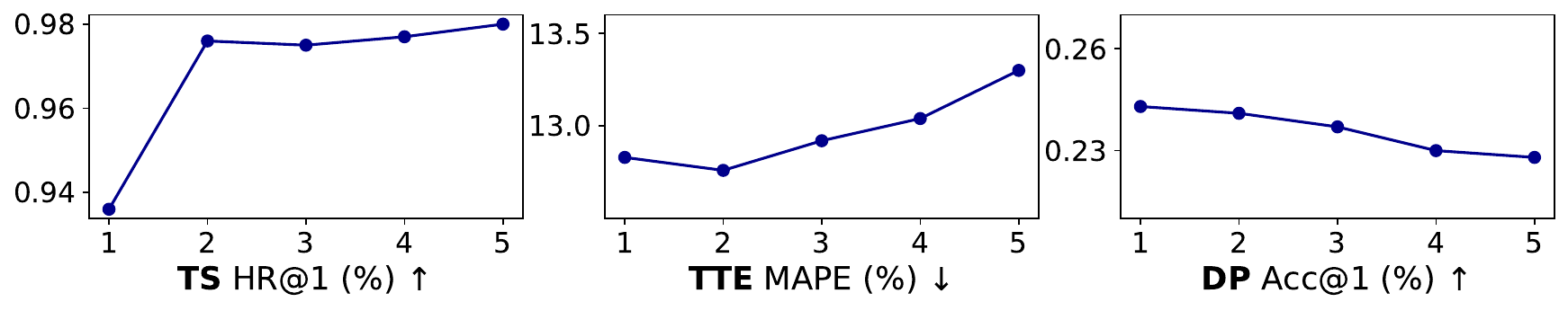}}
    \vspace{10pt}
    
    \subfloat[Number of negative samples $|\mathcal{Q}_{neg}|$.]
    {\includegraphics[width=0.5\textwidth]{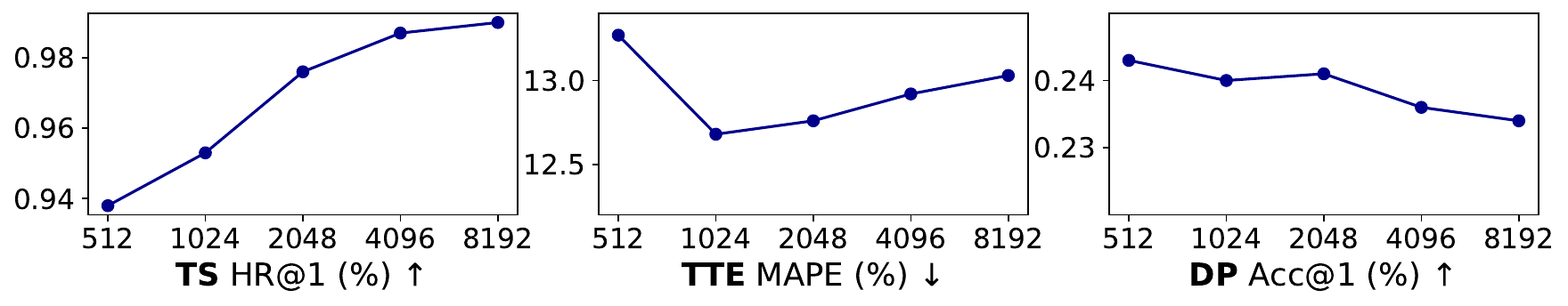}}
    \vspace{10pt}
    
    \subfloat[Embedding dimension.]
    {\includegraphics[width=0.5\textwidth]{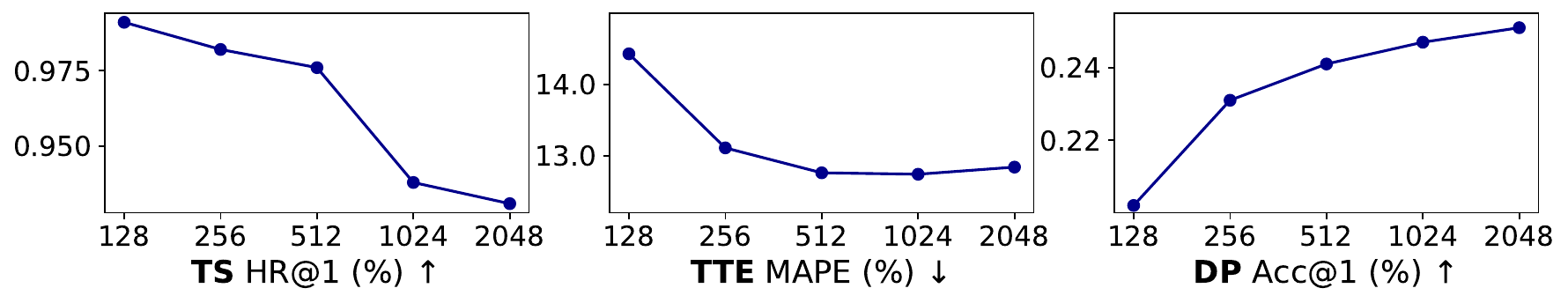}}
    \vspace{10pt}
    
    \subfloat[Masking ratio for $p_{RM}$, $p_{TC}$ and $p_{CM}$.]
    {\includegraphics[width=0.5\textwidth]{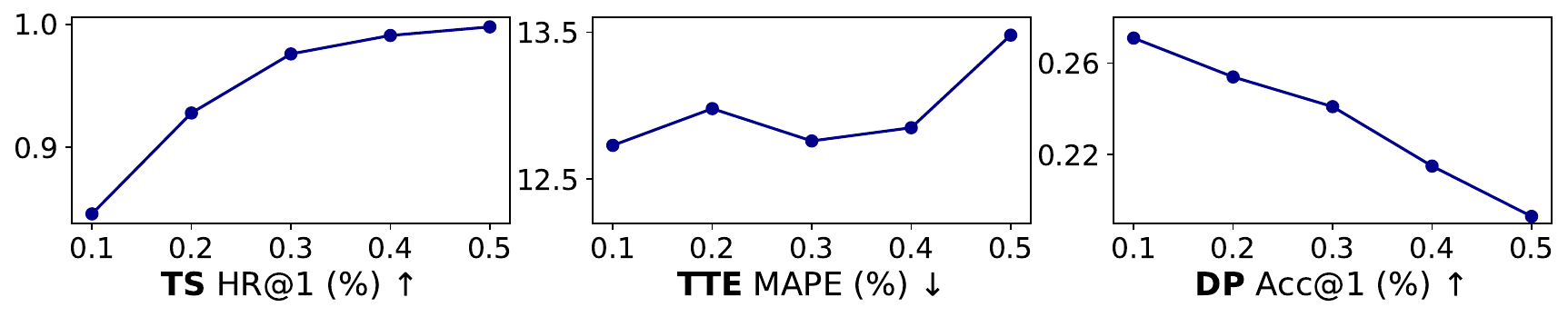}}
    
    \caption{Parameter analysis on the Porto dataset.}
    \label{fig:param_exp}
\end{figure}

We analyze the impact of four key hyperparameters on \approach{}'s performance across all downstream tasks, as demonstrated in Fig. \ref{fig:param_exp}:

\subsubsection{Encoder layers}
Performance across tasks shows minimal variation with encoder depth. 
For \textbf{TS}, performance peaks at two layers and remains stable when adding more layers. \textbf{TTE} and \textbf{DP} performance slightly decreases with deeper architectures. This suggests that a shallow architecture of 2 layers is sufficient to capture relevant trajectory characteristics.

\subsubsection{Negative sample queue size $|\mathcal{Q}_{neg}|$}
$|\mathcal{Q}_{neg}|$ primarily influences \textbf{TS} performance, as more negative samples help the method to contrast similar trajectories to dissimilar ones, thus reducing bias. 
For \textbf{TTE} and \textbf{DP}, the performance decreases slightly after a negative queue size of 1024.
Considering the trade-off between performance and training efficiency, we set $\mathcal{Q}_{neg} = 2048$.

\subsubsection{Embedding dimension} 
The embedding dimension affects \textbf{DP} most significantly, showing a 24.3\% improvement from lowest to highest dimension, indicating benefits from higher capacity. \textbf{TS} decreases with increasing dimensions, dropping notably at 1024, suggesting smaller embedding spaces better capture trajectory similarities. \textbf{TTE} improves up to 512 dimensions before plateauing. We choose 512 dimensions as the optimal tradeoff across tasks.

\subsubsection{Masking ratio} 
The masking ratio shows significant task-dependent effects.
For \textbf{TS}, performance improves with higher masking ratios, peaking near 1.0, indicating that strong masking enables learning robust trajectory representations.
\textbf{TTE} remains stable across ratios but declines at the highest ratio of 0.5.
\textbf{DP} performance gradually decreases with increased masking, suggesting benefits from fine-grained trajectory information.
We select a masking ratio of 0.3 for $p_{RM}$, $p_{TC}$ and $p_{CM}$ to balance performance across tasks.

An interesting finding from our parameter analysis is the inverse correlation between \textbf{TS} and \textbf{DP} performance across different parameters and their settings, particularly evident in the masking ratio and embedding dimension analysis. 
Higher embedding dimensions and lower masking ratios possibly preserve more precise, location-specific attributes, benefiting \textbf{DP}. 
In contrast, lower dimensions and higher masking ratios may facilitate the extraction of abstract route-level patterns, that capture overall trajectory patterns and thus benefit \textbf{TS}. 
Based on our analysis, we select two encoder layers, 2048 negative samples, 512-dimensional embeddings, and a 0.3 masking ratio as our default configuration.

\section{Related Work}
\label{sec:RelatedWork}

Trajectory Representation Learning encodes raw spatial-temporal data into a compact feature vector, capturing essential trajectory characteristics. This simplifies data analysis, enables efficient similarity search, and removes the need for manual feature engineering and task-specific models. 
TRL methods can be categorized into \textit{grid-based} and \textit{road-based} methods.

\textbf{Grid-based TRL}
A grid divides the spatial domain into a regular grid, which preserves structural and spatial concepts and thus is used in many spatio\--temporal tasks, such as traffic prediction \cite{metacitta}, spatial representation learning \cite{space2vec} and remote sensing \cite{tile2vec, simpoi1}.
T2vec \cite{t2vec} downsampled and distorted input trajectories and reconstructed the original trajectories utilizing a seq2seq model to obtain robust trajectory representations. 
CLT-Sim \cite{cltsim} utilized the same idea as t2vec and introduced contrastive learning to guide the model by using the pair of downsampled and distorted trajectory and the original trajectory as positives.
CSTTE \cite{cstte} enriches cell embeddings by adding positional encodings of time and location and utilizes a transformer-based encoder to model the long-term spatio-temporal correlations. 
TrajCL \cite{trajcl} introduced a dual-feature transformer-based trajectory encoder utilizing both cell embeddings and extracted location features.

\textbf{Road-based TRL}
Map matching algorithms allow trajectories to be represented as sequences of road segments in a road network \cite{mapmatchingsurvey1}. 
Additionally, road network representation learning \cite{hrnr, rn2vec} facilitates leveraging the road network topology and semantics to learn more meaningful trajectory representations.
Trembr \cite{trembr} utilizes an RNN-based encoder-decoder model and reconstructs the input trajectory.
PIM \cite{PIM} captures local and global information leveraging curriculum learning.
Toast \cite{toast} leverages a transformer-based encoder and employs masked token prediction by masking and then predicting road segments. 
JCLRNT \cite{jclm} learns road network and trajectory representations jointly by proposing a road-road, trajectory-trajectory, and road-trajectory contrast.
MMTEC \cite{mmtec} and JGRM \cite{jgrm} leverage GPS points to extract more detailed trajectory characteristics.
START \cite{start} incorporates temporal regularities and combines the masked token prediction objective with contrastive learning.  
LightPath \cite{lightpath} proposes a lightweight TRL method by utilizing sparse paths to reduce model complexity.

In contrast to these related works, \approach{} combines both modalities to facilitate a more intricate extraction of various trajectory characteristics. 
Further, while previous works modeled the road network as static, \approach{} incorporates the extraction of spatio-temporal dynamics, enabling the model to learn traffic patterns critical to urban mobility.

\section{Conclusion}
\label{sec:Conclusion}

In this paper, we introduced \approach{}, a novel trajectory representation learning model that effectively integrates road and grid modalities while incorporating spatio-temporal dynamics. 
Our novel spatio-temporal extraction method enables \approach{} to capture dynamic traffic patterns and temporal regularities. 
Further, we propose a three-branch architecture to enable \approach{} to process grid, road network, and spatio-temporal dynamics in parallel. 
Our inter and intra-contrastive losses align representations both across and within branches.
%
Through extensive experimentation on two real-world datasets, we demonstrated that \approach{} substantially outperforms state-of-the-art methods
across three downstream tasks, i.e., up to 43.22\% for trajectory similarity, up to 16.65\% for travel time estimation, and up to 10.16\% for destination prediction. 
%
Moreover, we conducted the first comprehensive comparison between grid-based and road-based trajectory representation learning methods, revealing that grid-based approaches excel in travel time estimation while road-based methods show advantages in trajectory similarity computation.
These findings emphasize the potential of combining grid and road modalities and 
the importance of 
capturing dynamic spatio-temporal patterns of urban traffic.

\bibliographystyle{IEEEtran}
\bibliography{references}

\end{document}